\def\year{2019}\relax
\documentclass[letterpaper]{article} 
\usepackage{aaai19}  
\usepackage{times}  
\usepackage{helvet}  
\usepackage{courier}  
\usepackage{url}  
\usepackage{graphicx}  
\usepackage{dirtytalk}
\usepackage{graphicx}
\usepackage{times}
\usepackage{latexsym}
\usepackage{svg}
\usepackage{amsmath}
\usepackage{url}
\usepackage{array}
\usepackage{balance}
\usepackage{pifont}
\usepackage{caption}
\usepackage{amssymb}
\usepackage{multirow}
\usepackage{svg}
\usepackage{fontawesome}
\usepackage{algorithm}
\usepackage[noend]{algpseudocode}
\usepackage{textcomp}
\floatname{algorithm}{Algorithm}

\newcolumntype{L}[1]{>{\raggedright\let\newline\\\arraybackslash\hspace{0pt}}m{#1}}
\newcolumntype{C}[1]{>{\centering\let\newline\\\arraybackslash\hspace{0pt}}m{#1}}
\newcolumntype{R}[1]{>{\raggedleft\let\newline\\\arraybackslash\hspace{0pt}}m{#1}}

\begin{document}
\title{Transfer Learning for Sequence Labeling Using Source Model and Target Data}
\author{Lingzhen Chen \\
  University of Trento \\
  Povo, Italy \\
  {\tt lingzhen.chen@unitn.it} \\\And
  Alessandro Moschitti\thanks{The main part of this work was carried out when the author was at the Unviversity of Trento. } \\
  Amazon \\
  Manhattan Beach, CA, USA \\
  {\tt amosch@amazon.com} \\}

\maketitle
\begin{abstract}
In this paper, we propose an approach for transferring the knowledge of a neural model for sequence labeling, learned from the source domain, to a new model trained on a target domain, where new label categories appear. 
Our transfer learning (TL) techniques enable to adapt the source model using the target data and new categories, without accessing to the source data. 
Our solution consists in adding new neurons in the output layer of the target model and transferring parameters from the source model, which are then fine-tuned with the target data. 
Additionally, we propose a neural adapter to learn the difference between the source and the target label distribution, which provides additional important information to the target model.
Our experiments on Named Entity Recognition show that (i) the learned knowledge in the source model can be effectively transferred when the target data contains new categories and (ii) our neural adapter further improves such transfer.
\end{abstract}

\section{Introduction}
One important challenge of sequence labeling tasks concerns dealing with the change of the application domain during time. For example, in standard concept segmentation and labeling \cite{wang2005spoken}, semantic categories, e.g., \emph{departure} or \emph{arrival} cities, vary according to new scenarios, e.g., \emph{low-cost flight} or \emph{budget terminal} were not available when the  Automatic Terminal Information Service (ATIS) corpus was compiled \cite{Hemphill:1990:ASL:116580.116613}. Similar rationale applies to another very important sequence labeling task, Named Entity Recognition (NER), where entities in a domain are continuously evolving, e.g., see the \emph{smart phone} domain. 

Standard models for sequence labeling are supposed to be trained and applied to the data with the same set of categories, which may limit the reuse of previous models.
As an example for NER, users interested in finance would probably target entities such as Companies or Banks while other users interested in politics want to recognize Senators, Bills, Ministries, etc. Besides these domain-specific NEs, there may be common categories, such as Location or Date. Hence, if there are models pretrained on data with these common NEs, it would be useful to have some methods that modify them to serve customized NER purposes.
Even in the same application domains, NE categories can vary over time due to, for example, the introduction of new products.  

In these cases, a typical sequence labeling approach would be to (i) build a new dataset including the annotation of the new categories; and (ii) retrain a model from scratch. 
However, in addition to the disadvantage of retraining models and re-annotating all the documents, some industrial scenarios prevent the release of training data (e.g., for copyright or privacy concerns) to rebuild the models.  

Motivated by the problems described above, in this work, we define a new paradigm for a progressive learning of sequence labeling tasks.  To simplify our study, without loss of generality, we define a setting consisting of two aspects:
(i)  a source model, $\mathcal{M}_{\text{S}}$, already trained to recognize a certain number of categories on the source data, $\mathcal{D}_{\text{S}}$; and
(ii) a TL task consisting in training a new model, $\mathcal{M}_{\text{T}}$, on the target data, $\mathcal{D}_{\text{T}}$,  where new categories appear, in addition to those of the $\mathcal{D}_{\text{S}}$ (note that $\mathcal{D}_{\text{S}}$ is no longer available to perform TL). $\mathcal{D}_{\text{T}}$ is typically much smaller in size compared to $\mathcal{D}_{\text{S}}$. 
 These kinds of problems regard leveraging knowledge about a model learned on a source domain, to improve learning a model for another task on a target domain \cite{PanY10}. One variation of TL is a setting where the target domain does not change while the output space of the target task changes. This corresponds to our described progressive sequence labeling setting.

To tackle the problem, we propose two neural methods of TL for progressive labelling. 
Firstly, given an initial neural model  $\mathcal{M}_{\text{S}}$ trained on source data $\mathcal{D}_{\text{S}}$, we modify its output layer to include new neurons for learning the new categories and then we continue to train on $\mathcal{D}_{\text{T}}$.
More specifically, we implement a Bidirectional LSTM (BLSTM) with Conditional Random Fields (CRF) as $\mathcal{M}_{\text{S}}$. In the transfer learning step, we modify such architecture to build $\mathcal{M}_{\text{T}}$, reuse previous weights and fine-tuning them on $\mathcal{D}_{\text{T}}$, where again the assumption is that $\mathcal{D}_{\text{T}}$ contains both seen and unseen categories.

Secondly, as a refinement of the first approach, we propose to use a neural adapter: it connects $\mathcal{M}_{\text{S}}$ to $\mathcal{M}_{\text{T}}$, also enabling the latter to use the features from the former.
The connection is realized by a BLSTM, which takes the hidden activations in $\mathcal{M}_{\text{S}}$ as an additional input to $\mathcal{M}_{\text{T}}$.
Besides better utilizing the learned knowledge, the aim of the neural adapter is to mitigate the effect of label disagreement, e.g., in some cases, the surface form of a new category type has already appeared in the $\mathcal{D}_{\text{S}}$, but they are not annotated as a label. Because it is not yet considered as a concept to be recognized.
Note that the parameters of $\mathcal{M}_{\text{S}}$ is not updated during training $\mathcal{M}_{\text{T}}$. 

Our models and procedures apply to any sequence labeling task,
however, to effectively demonstrate the impact of our approach, also considering the space available in this paper, we focus on NER.
We analyze the performance of both our methods by testing the transfer of different categories. Our main contribution is therefore twofold:
firstly, we show that our pre-training and fine-tuning method can utilize well the learned knowledge for the target labeling task while being able to learn new knowledge.

Secondly, we show that our proposed neural adapter has the ability to mitigate the forgetting of previously learned knowledge, to combat the annotation disagreement, and to further improve the transferred model performance. 

Finally, to claim that our approach works for general sequence labeling would require testing it on different tasks and domains. However, we observe that our approach only exploits the concept of category sequence, where the words compounding such sequences (boundaries) are annotated with a standard BIO (Begin, In, Other) tagging. Thus, there is no apparent reason to prevent the use of our approach with different tasks and domains.
We made our source code and the exact partition of our dataset available for  further research. \footnote{\url{https://github.com/liah-chan/transferNER}}

\section{Related Work}
\paragraph{Named Entity Recognition}
In the earlier years of NER, most work approached the task by engineering linguistic features \cite{ChieuN03,CarrerasMP03a}. 
Machine learning algorithms such as Maximum Entropy, Perceptron and CRFs were typically applied \cite{FlorianIJ003,ChieuN03,DiesnerC08,HeK08}.

 Recent work mainly includes neural models, where the current state of the art is given by Recurrent Neural Network models, which incorporate word and character level embeddings and/or additional morphological features.
 Huang, Xu, and Yu, \shortcite{HuangXY15} uses BLSTM combined with CRF to established the state-of-the-art performance on 
 NER (90.10 in terms of test F1 on CONLL 2003 NER dataset). 
 Later, Lample et al., \shortcite{LampleBSKD16} implemented the same CRF over BLSTM model without using any handcraft features. They reported 90.94 of test F1 on the same dataset. 
Chiu and Nichols, \shortcite{ChiuN16} also implemented a similar BLSTM model with Convolutional filters as character feature extractor, achieving 91.62 in the F1 score (BLSTM$+$CNN$+$lexical features).
 
 In this work, we opt to use the BLSTM and BLSTM + CRF for NER with Transfer Learning, in order to test whether our proposed methods can be applied on the state-of-the-art models.
\paragraph{Transfer Learning}
 Neural networks based TL has proven to be very effective for image recognition ~\cite{DonahueJVHZTD14,RazavianASC14}. 
 As for NLP, Mou et al.,~\shortcite{MouMYLX0J16} showed that TL can also be successfully applied on semantically equivalent NLP tasks.
 Researches were carried out on NER related TL too.
Qu et al., \shortcite{QuFZHB16} explored TL for NER with different NE categories (different output spaces). 
 They pre-train a linear-chain CRF on large amount annotated data in the source domain. A two linear layer neural network to learn the discrepancy between the source and target label distributions. Finally, they initialize another CRF with learned weight parameters in linear layers for the target domain. 
Kim et al., \shortcite{KimSSJ15} experimented with transferring features and model parameters between similar domains, where the label types are different but may have semantic similarity. Their main approach is to construct label embeddings to automatically map the source and target label types to help improve the transfer.

 In our work, we aim to transfer knowledge in an incremental, progressive way within the same domain, rather than to other domains. We assume that the target output space includes the source output space.  In terms of mitigating the discrepancies between the source and target label distribution, we propose a neural adapter to learn them.

In Rusu et al.,\shortcite{RusuRDSKKPH16} also used an adapter to help transfer. 
 They proposed progressive networks that solve sequence of reinforcement learning tasks while being immune to  parameter forgetting. The networks leverage knowledge learned with an adapter, which is an additional connection between new model and learned models.  This connection is realized by a feed-forward neural layer with non-linear activation.
 Due to the characteristics of sequence labeling tasks, we proposed to use BLSTM in a sequence-to-sequence way that learns to map the output sequence in the source space to the output sequence in the target space.

\section{State-of-the-art in Neural Sequence Labeling}
A standard sequence labeing problem can be defined as follow: given an input sequence $X = x_1, x_2, ..., x_n$ $(x_i \in \mathcal{X})$,  predict the output sequence $Y = y_1, y_2, ..., y_n$ $ (y_i \in \mathcal{Y})$.
$\mathcal{X}$ and $\mathcal{Y}$ represent the input and output space respectively.
Typically, the model learns to maximize the conditional probability $P(Y|X)$.

In this section, we introduce two state-of-the-art neural models for learning $P(Y|X)$, i.e., BLSTM and BLSTM$+$CRF, which are also the base models we use to progressively learn Named Entities. Note that such approaches are the state of the art in case of NER, thus we study the effectiveness of our proposed transfer learning method in a state-of-the-art setting. The general architecture is described on the left side of Figure \ref{fig:architec}. This is composed of: a BLSTM at character level, followed by a BLSTM at word level, a fully connected layer and a CRF/output layer. The individual components are described in the next sections.

\subsection{Word \& Character Embeddings}
A word in the input sequence is represented by both its word-level and character-level embeddings. 
We use pretrained word embeddings to initialize a lookup table to map the input word $x$ (represented by an integer index) to a vector $\boldsymbol{w}$.
A character-level representation is typically used because the NER task is sensitive to the morphological traits of a word such as capitalization.
They were shown to provide useful information for NER \cite{LampleBSKD16}.
 We use a randomly initialized character embedding lookup table and then pass the embeddings to a BLSTM  to obtain character level embedding $\boldsymbol{e}$ for $x$ (the details are described in the following section).
The final representation of the $t$th word $x_t$ in the input sequence is the concatenation of its word-level embedding $\boldsymbol{w_t}$ and character-level embedding $\boldsymbol{e_t}$.

\subsection{Bidirectional LSTM}
\label{ssec:blstm}

BLSTM is composed of a forward LSTM ($\overrightarrow{\text{LSTM}}$) and a backward LSTM ($\overleftarrow{\text{LSTM}}$), which read the input sequence (represented as word vectors described in the previous subsection) in both left-to-right and reverse order. 
The output of the BLSTM $\boldsymbol{h}_t$ is obtained by the concatenation of forward and backward output: $\boldsymbol{h}_t = [\overrightarrow{\boldsymbol{h}_t}; \overleftarrow{\boldsymbol{h}_t}]$, where
$\overrightarrow{\boldsymbol{h}_t} = \overrightarrow{\text{LSTM}}(\boldsymbol{x}_t, \overrightarrow{\boldsymbol{h}}_{t-1})
$ and 
$
\overleftarrow{\boldsymbol{h}_t} = \overleftarrow{\text{LSTM}}(\boldsymbol{x}_t, \overleftarrow{\boldsymbol{h}}_{t+1})$. 
$\boldsymbol{h}_t$ captures the left and right context for $\boldsymbol{x}_t$ and is then passed to a fully-connected layer, $\boldsymbol{p}_t$. 
The final prediction $\boldsymbol{y}_t$ is obtained applying a softmax over the output layer, i.e.,
$$\text{P}(y_t = c | p_t) = \frac{e^{W_{O, c} \cdot p_t}}{\sum_{c' \in C} e^{W_{O,c'} \cdot p_t}},
$$
where $W_O$ are parameters to be learned on the output layer and $C$ represents the set of all the possible output labels.

In the case of the character-level BSLTM, the forward and backward LSTMs take the sequence of character vectors $[\boldsymbol{z}_1, \boldsymbol{z}_2, ..., \boldsymbol{z}_k]$ as input, where $k$ is the number of characters in a word. 
The final character level embedding $\boldsymbol{e}_t$ for word $x_t$ is obtained by concatenating $\overrightarrow{\boldsymbol{e}_t}$ and $\overleftarrow{\boldsymbol{e}_t}$.

\subsection{CRF Tagging}
We implement a Linear Chain CRF \cite{LaffertyMP01} model over BLSTM to improve the prediction ability of the model, by taking the neighboring prediction into account while making the current prediction. 
Here, $\text{P}(Y|X)$ is computed by
$
     \text{P}(Y|X) = \frac{
    e^{s(X,Y)}
    }{
    \sum_{Y' \in \mathcal{Y}} e^{s(X,Y')}
    }
$, 
 where $\mathcal{Y}$ is all possible label sequences and $s(\cdot,\cdot)$ is calculated by adding up the transition and emission scores for a label sequence.
 To be more specific, the emission score is the probability of predicting the label $y_t$ for $t$th word in the sequence. The transition score is the probability of transiting from previously predicted label $y_{t-1}$ to current label $y_t$.
 The outputs of fully-connected layer $\boldsymbol{p}_t$ at time step $t$ provides the emission scores for all possible value of $y$. 
 A square matrix $\textbf{P}$ of size $C+2$ is used to store transitional probabilities among $C$ output labels, as well as a \textit{start} label and an \textit{end} label.
 Hence,  
 \begin{equation*}
     s(X,Y) = \sum_t \boldsymbol{p}_t[y_t] +
    \textbf{P}_{y_t, y_{t-1}}
 \end{equation*}

\section{Our Progressive Adaptation Models}
In this section, we formalize our progressive learning problem and describe our proposed TL method in detail.

\begin{figure}[t]
  \includegraphics[width=0.48\textwidth]{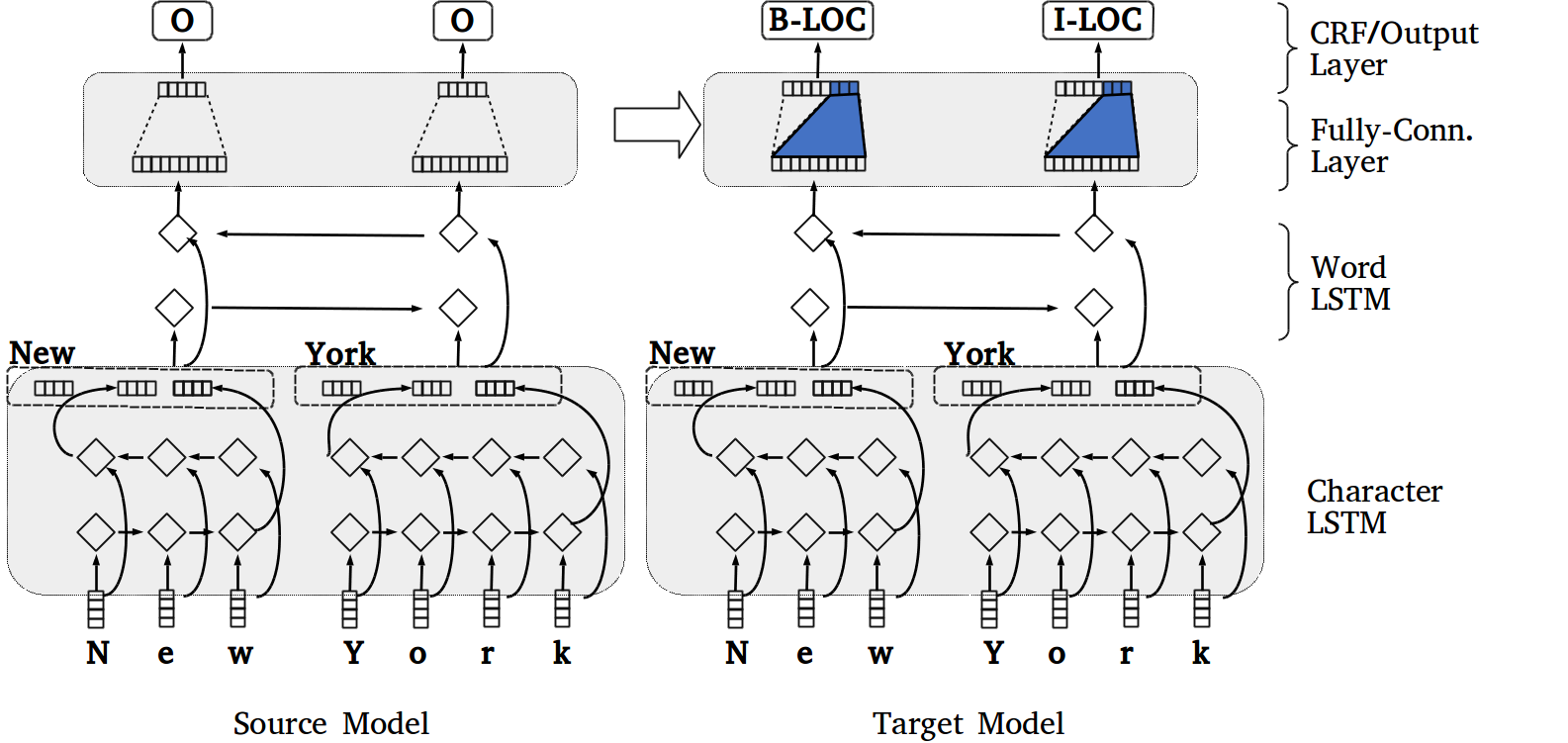}
  \caption{\small Source and target model architecture}
  \label{fig:architec}
\end{figure}

\subsection{Problem Formalization}
In the initial phase, a sequence labeling model, $\mathcal{M}_{\text{S}}$, is trained on a source dataset, $\mathcal{D}_{\text{S}}$, which has $E$ classes. Then, in the next phase, a new model, $\mathcal{M}_{\text{T}}$, needs to be learned on target dataset, $\mathcal{D}_{\text{T}}$, which contains new input examples and $E+M$ classes, where $M$ is the number of new classes. 
$\mathcal{D}_{\text{S}}$ cannot be used for training  $\mathcal{M}_{\text{T}}$.
\subsection{Our Transfer Learning Approach}
Given  pre-trained $\mathcal{M}_{\text{S}}$ model, our first proposed method to progressively recognize new categories consists in transferring parameters to $\mathcal{M}_{\text{T}}$ and then fine-tuning it.
 
 \begin{algorithm}
\small
\caption{Source Model Training}
\label{alg:init-step}
\begin{algorithmic}[1]
\Require 
$\{(X^{(n)}, Y^{(n)})\}_{n=1}^{N_{\text{S}}}$: source training data.

$\hspace{.6em} \{(X^{(n)}, Y^{(n)})\}_{n=N_{\text{S}}+1}^{N_{\text{V}}}$: validation data.
    
$\hspace{.6em} \mathcal{L}$: loss function.

$\hspace{.6em}tp=0$: temporary variable (best valid. F1).   

$\hspace{.6em}\mathcal{F}$: evaluation function.

\Ensure  $\{y^{(n)}\}_{n=1}^{N_{\text{S}}}$: predictions for training data

$\{y^{(n)}\}_{n=N_\text{S}+1}^{N_{\text{V}}}$: predictions for validation data

 $\hspace{.6em}\hat{\theta}^{\text{S}}$: Optimal parameters for source model
\State Randomly initialize $\theta^{\text{S}}$
\For{$e = 1 \to n\_epochs$}
\For{$n = 1 \to N_{\text{S}}$} \Comment{training step}
  \State $y^{(n)} = \mathcal{M}(X^{(n)})$
  \State $\theta^{\text{S}} := \theta^{\text{S}} - \alpha \Delta_{\theta^{\text{S}}} \mathcal{L}[y^{(n)}, \theta^{\text{S}}; X^{(n)},Y^{(n)}]$
\EndFor
\For{$n = N_{\text{S}}+1 \to N_{\text{V}}}$ \Comment{predictions over the valid.~set}
  \State $y^{(n)} = \mathcal{M}(X^{(n)})$
\EndFor
  \If { $\mathcal{F}(\{y^{(n)}\}_{n=N_{\text{S}}+1}^{N_{\text{V}}}) > tp$} \Comment{F1 over the valid.~set}
  	\State  $\hat{\theta}^{\text{S}} := \theta^{\text{S}}$ ; $tp=\mathcal{F}(\{y^{(n)}\}_{n=N_{\text{S}+1}}^{N_{\text{V}}})$
  \State Save  $\hat{\theta}^{\text{S}}$
  \EndIf
\EndFor
        \end{algorithmic}
    \end{algorithm}

\subsubsection{Training of a source model}
We supposed that a sequence labeling model is trained on source data until the optimal parameters $\hat{\theta}^{\text{S}}$ are obtained.  These will be saved and reused for transfer learning. 
The details of such training are illustrated by Algorithm~\ref{alg:init-step}. This takes (i) $N^{\text{S}}$ training examples in the source data and (ii) uses Multi-class Cross-Entropy and the F1 score as the loss $\mathcal{L}$ and evaluation function $\mathcal{F}$, respectively. 
The predictions $y^{(n)}$ on the input $X^{(n)}$ are obtained by forward propagation through the model $\mathcal{M}$. 
The parameters are updated by a learning rate $\alpha$. The final model corresponds to the highest evaluation metric $\mathcal{F}(\{y^{(n)}\}_{n=N_{\text{S}+1}}^{N_{\text{V}}})$ computed on the validation set in $n\_epochs$.

\subsubsection{Parameter Transfer}
To enable the recognition of a new category, we modify the fully-connected layer after BLSTM and the output layer of the network. The right side of Figure~\ref{fig:architec} shows the difference between the source model (on the left side) and our transferred model in blue color.
In more detail, the fully-connected layer after the word BLSTM maps the output $\boldsymbol{h}$ to a vector $\boldsymbol{p}$ of size $nE$. $n$ is a factor depending on the tagging format of the dataset (e.g., $n=2$ if the dataset is in BIO format, since for each NE category, there would be two output labels \texttt{B-NE} and \texttt{I-NE}). 
Therefore, we extend the output layer by size $nM$, where $M$ is the number of new categories.

\begin{algorithm}
\small
\caption{Parameter Transfer}
\label{alg:inter-step}
\begin{algorithmic}[1]
\Require $\hat{\theta}^{\text{S}}$: optimal parameters for the source model 

\Ensure $\theta^{\text{T}}$: initial parameters of the target model  

\For{$\theta_O$ in $W_O$} \Comment{parameters in the output layer}
\State $\theta_O$ := ReInit() \Comment{draw from normal distribution}
\EndFor

\For{$\theta_{\bar O}$ in $W_{{\bar O}}$} \Comment{parameters in other layers}
\State $\theta_{\bar O} := \hat{\theta}^{\text{S}}_{\bar{O}}$ \Comment{copy from trained parameter}
\EndFor 
\State $\theta^{\text{T}} = (\theta_O, \theta_{{\bar O}}) $
\State \Return{$\theta^{\text{T}}$}
\end{algorithmic}
\end{algorithm}

The extended part above, i.e., parameters in the output layer, ${\theta}_{O}$, is initialized with weights drawn from the normal distribution, $ X \sim \mathcal{N}(\mu, \sigma^{2})$, where $\mu$ and $\sigma$ are the mean and standard deviation of the pre-trained weights in the same layer. This is denoted by ReInit() in Algorithm~\ref{alg:inter-step} of the Parameter Transfer step.

In contrast, all the other parameters, ${\theta}_{\bar{O}}$, i.e, those not in the output layer, are initialized with the corresponding parameters from the source model, i.e., $\hat{\theta}^{\text{S}}_{\bar{O}}$.
%
 This way, the associated weight matrix of the fully-connected layer $\textbf{W}^s_O$ also updates from the original shape $nC\times p$ to a new matrix $\textbf{W}^{'s}_O $ of shape $(nC+nM)\times p$. 
 Note that the parameters $\hat{\theta}^{\text{S}}_{O}$ and all the other parameters, $\hat{\theta}^{\text{S}}_{\bar{O}}$, are essentially the weights in the matrix $\textbf{W}^s_O$ and the weights in the other layers.
   

\begin{algorithm}
\small
\caption{Target Model Training}
\label{alg:sub-step}
\begin{algorithmic}[1]
\Require $\{(X^{(n)}, Y^{(n)})\}_{n=1}^{N_{\text{T}}}$: target training data.
    
$\mathcal{L}$: loss function.
    

$\mathcal{F}$: evaluation function

$\theta^{\text{T}}$: transferred parameters
\Ensure $\{y^{(n)}\}_{n=1}^{N_{\text{T}}}$: predictions
\For{$e = 1 \to n\_epochs$}
\For{$n = 1 \to N_T$}
  \State $y^{(n)} = \mathcal{M}(X^{(n)})$
  \State $\theta^{\text{T}} := \theta^{\text{T}} - \alpha \Delta_{\theta^{\text{T}}} \mathcal{L}[y^{(n)},\theta^{\text{T}}; X^{(n)},Y^{(n)}]$
\EndFor
\EndFor{}
        \end{algorithmic}
    \end{algorithm}
\subsubsection{Training the target model}
In Algorithm~\ref{alg:sub-step}, the new parameters are updated as a standard training cycle (we use a validation set and early stopping as in the source model training, for brevity we did not put this step in the Algorithm~\ref{alg:sub-step}). To update the BLSTM$+$CRF model, we also modify the transitional matrix to include the transition probability among new labels and other previously seen labels. 

\begin{figure}[t]
  \centering
  \includegraphics[width=0.4\textwidth]{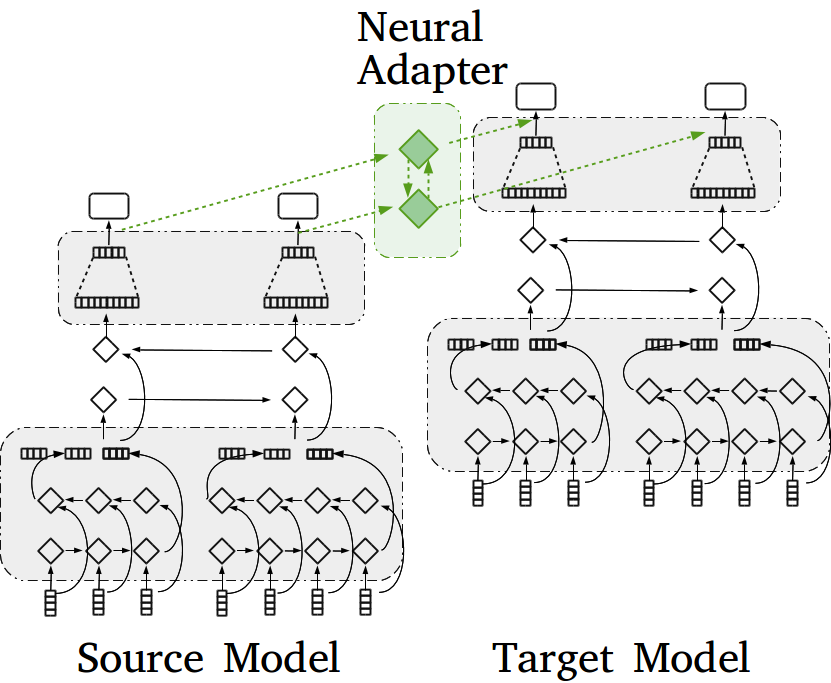}
  \caption{\small \label{fig:adapter} Our Proposed Neural Adapter}
\end{figure}
\subsection{Transfer Learning using neural adapters}
It should be noted that many word sequences corresponding to new NE categories can already appear in the source data, but they are annotated as \texttt{null} since their label is not part of the source data annotation yet.
This is a critical aspect to solve as otherwise the target model with transferred parameters would treat the word sequence corresponding to a new NE category as a null category.

We design a neural adapter, shown in Fig.~\ref{fig:adapter}, to solve the problem of disagreement in annotations between the source and target data.  This is essentially a component that helps to map the predictions from the output space of the source domain into that of the target domain. 
The figure shows that the target model $\mathcal{M}_{\text{T}}$ gets the hidden activation of the last layer in $\mathcal{M}_{\text{S}}$ as an additional input. Basically, the neural adapter connects each output of the fully-connected layer in $\mathcal{M}_{\text{S}}$ to the corresponding output of $\mathcal{M}_{\text{T}}$. 
 
 More precisely, we use $\overrightarrow{\text{A}}$ and $\overleftarrow{\text{A}}$ to denote the forward and backward adapter (i.e., a BLSTM).
  It takes the output of the fully-connected layer $\boldsymbol{p}_t^{\text{S}}$ as input at each time step $t$.
  The output of $\overrightarrow{\text{A}}$ and $\overleftarrow{\text{A}}$ are computed as $\overrightarrow{\boldsymbol{a}}_t =
  \overrightarrow{\text{A}}(\boldsymbol{p}_t^{\text{S}}, 
  \overrightarrow{\boldsymbol{a}}_{t-1})$ 
  and $\overleftarrow{\boldsymbol{a}}_t = \overleftarrow{\text{A}}(\boldsymbol{p}_t^{\text{S}}, \overleftarrow{\boldsymbol{a}}_{t+1})$, respectively.
 The output of the target model consists in a softmax over the output of the fully-connected layer  $\boldsymbol{p}_t^{\text{T}\textquotesingle}$
 obtained by
$ \boldsymbol{p}_t^{\text{T}\textquotesingle} = \boldsymbol{a}_t
  \oplus \boldsymbol{p}_t^{\text{\text{T}}}$,
  where $\boldsymbol{a}_t = [\overrightarrow{\boldsymbol{a}}_t \oplus \overleftarrow{\boldsymbol{a}}_t]$ and $\oplus$ is the element-wise summation.
 The parameters of the adapter are jointly learned in the subsequent step with the rest of the target model parameters. The parameters of the source model is, however, not updated. 
 
The choice of BLSTM as the adapter is motivated by the fact that we want to incorporate the context information of a feature in the sequence to detect the new category that was annotated and possibly incorrectly predicted as not a label.

\section{Experiments}
Although our approach is generally valid for any sequence labeling task, in these experiments, we focus on NER.
We test our basic transfer approach, then we show the impact of our proposed neural adapter, testing several transfer learning options on just one category, i.e., LOC. 
Finally, we provide more general results of the best model, applying it to different settings/data obtained by selecting one category as the new target category (among to the available ones).
\subsection{Datasets}
We primarily used CONLL 2003 NER \footnote{https://www.clips.uantwerpen.be/conll2003/ner/} dataset for our experiments. We modified it to simulate our progressive learning task.
The original dataset includes news articles with four types of named entities -- organization, person, location and miscellaneous (represented by \texttt{ORG}, \texttt{PER}, \texttt{LOC}, and \texttt{MISC}, respectively). 
For the purpose of our experiment, we divide the CONLL train set in 80\%/20\% as $\mathcal{D}_{\text{S}}$ and $\mathcal{D}_{\text{T}}$, for the initial and subsequent steps of the experiments. Please note that in the subsequent step, $\mathcal{D}_{\text{S}}$ is no longer available.
We make \texttt{LOC} the new label to be detected in the subsequent step. 
Hence we replace all the \texttt{LOC} label annotations with \texttt{null}, when they appear in $\mathcal{D}_{\text{S}}$ (the validation and test subsets). Instead,  we keep \texttt{LOC} as it is in $\mathcal{D}_{\text{T}}$. We repeat this process for all four categories to obtain four datasets for our TL setting. 

To demonstrate that our method can be applied independently of the language, we also carry out experiments on I-CAB (Italian Content Annotation Bank).\footnote{http://ontotext.fbk.eu/icab.html, used for EVALITA 2007 NER shared task}
 It does not come with a test set, hence we held out 30\% of training set as the \emph{official} test set and then carry out the same pre-processing as that for CONLL dataset. Hence, we obtain another four datasets for our setting.
 A summary of label statistics of these two datasets is shown in Table~\ref{tab:data-label}.
 
\begin{table}[t]
\small
\centering
\begin{tabular}{c|c|c|c|c|c}
\hline
\multicolumn{2}{c|}{\textbf{CONLL 03}}& \bf LOC & \bf PER & \bf ORG & \bf MISC \\ 
\hline
\multirow{2}{*}{Train}&$\mathcal{D}_{\text{S}}$ & 0 & 8948 & 8100 & 3686 \\
\cline{2-6}
&$\mathcal{D}_{\text{T}}$ & 1637 & 2180 & 1925 & 907 \\
\hline
\multicolumn{2}{c|}{Valid ($\mathcal{D}_{\text{S}}$ / $\mathcal{D}_{\text{T}}$)}& 0/2094 & 3146 & 2092 & 1268 \\
\hline
\multicolumn{2}{c|}{Test ($\mathcal{D}_{\text{S}}$ / $\mathcal{D}_{\text{T}}$)}& 0/1925 & 2773 & 2496 & 918 \\ 

\hline
\hline

\multicolumn{2}{c|}{\textbf{I-CAB}}& \bf GPE & \bf LOC & \bf PER &\bf ORG \\ 
\hline
\multirow{2}{*}{Train}&$\mathcal{D}_{\text{S}}$ & 0 & 247 & 2540 & 2471 \\
\cline{2-6}
&$\mathcal{D}_{\text{T}}$ & 310 & 78 & 676 & 621 \\
\hline
\multicolumn{2}{c|}{Valid ($\mathcal{D}_{\text{S}}$ / $\mathcal{D}_{\text{T}}$)}& 0/626 &174 & 1282 & 1302 \\
\hline
\multicolumn{2}{c|}{Test ($\mathcal{D}_{\text{S}}$ / $\mathcal{D}_{\text{T}}$)}& 0/1234 & 216 & 2503 & 1919 \\ 
\hline
\end{tabular}
\caption{\small Number of entities in CONLL dataset (in English) and I-CAB dataset (in Italian).}
\label{tab:data-label}
\end{table}

\subsection{Model selection and hyperparameters}
Apart from dividing the dataset into two sets as described in the previous section, we do not perform any specific pre-processing except for replacing all digits with 0 to reduce the size of the vocabulary.
We use 100 dimension \textsc{GloVe} pretrained embedding for English \footnote{http://nlp.stanford.edu/data/glove.6B.zip} and Italian \footnote{http://hlt.isti.cnr.it/wordembeddings/} to initialize the weights of the embedding layer. Since we do not lowercase the tokens in the input sequence, we map the words having no direct mapping to the pretrained word embeddings to their lowercased counterpart, if one is found in the pretrained word embeddings. 

We map the infrequent words (words that appear in the dataset for less than two) to \textit{$<$UNK$>$} as well as the unknown words appearing in the test set. The word embedding for \textit{$<$UNK$>$} is drawn from a uniform distribution between $[-0.25, 0.25]$.
The character embedding lookup table is randomly initialized with embedding size of 25.
The hidden size of the character-level BLSTM is 25 while the word level one is 128.

We apply a dropout regularization on the word embeddings with a rate of 0.5. 
All models are implemented in TensorFlow ~\cite{tensorflow}, as an extension of NeuroNER ~\cite{DernoncourtLS17}. We use Adam ~\cite{KingmaB14} optimizer with a learning rate of 0.001, gradient clipping of 50.0 to minimize the categorical cross entropy, and a 
maximum epoch number of 100 at each step. 
The models are evaluated with the F1 score as in the official CONLL 2003 shared task \cite{SangM03}.

\subsection{Results on CoNLL and I-CAB datasets}

As a first step, we verified that our implementations are at the state of the art by testing them on  traditional NER settings (i.e., the original CoNLL setting). Our BLSTM+CRF model achieved 91.26 and 80.59 (on I-CAB) in F1 without handcraft features. These results are comparable to the state-of-the-art performance on both dataset \cite{ChiuN16,MagniniCTBMLBCTLSS08}.

 \begin{table*}[!ht]
\centering
\small
\begin{tabular}
{c|c|C{0.9cm}|C{0.9cm}|C{0.9cm}|C{0.9cm}|C{0.9cm}|C{0.9cm}|C{0.9cm}|C{0.9cm}}
\hline

\multicolumn{2}{c|}{ } & \multicolumn{4}{c|}{CONLL 2003} & \multicolumn{4}{c}{I-CAB 2006} \\
\hline
\multirow{2}{*}{Model} & \multirow{2}{*}{Settings} & $\mathcal{M}_{\text{S}}$ & \multicolumn{3}{c|}{$\mathcal{M}_{\text{T}}$} & $\mathcal{M}_{\text{S}}$ & \multicolumn{3}{c}{$\mathcal{M}_{\text{T}}$} \\
\cline{3-10}
& & Ori & Ori. & New & All & Ori & Ori. & New & All \\ \hline
\multirow{5}{*}{BLSTM} & \faLock E \faLock B \faAdjust O & \multirow{5}{*}{91.06} & 90.41 & 86.08 & 89.39 & \multirow{5}{*}{74.78} & 73.95 & 10.31 & 60.58\\
& \faLock E \faLock B \faUnlock O && 90.04 & 84.94 & 88.83 &&  73.84 & 13.83 & 61.23 \\
& \faUnlock E \faUnlock B \faUnlock O && 90.42 & \textbf{89.39} & 90.18 && \bf 74.15 & 61.41 & 71.47 \\
& \faUnlock E \faUnlock B \faUnlock O \faExchange A && \bf 90.94  & 89.33  & \bf 90.56 && 74.12 & \bf 67.95 & \bf 72.82 \\
& \faSpinner E \faSpinner B \faSpinner O && 86.52 & 87.19 & 86.68 && 63.17 & 67.82 & 64.15 \\

\hline
\hline

\multirow{5}{*}{BLSTM$+$CRF} & \faLock E \faLock B \faAdjust O & \multirow{5}{*}{91.35}& 90.76 &45.89 & 80.11 & \multirow{5}{*}{76.86} & 69.62 & 52.52 & 69.62\\
& \faLock E \faLock B \faUnlock O && 90.20 & 68.66 & 85.09 && 74.43 & 41.80 & 67.57 \\
& \faUnlock E \faUnlock B \faUnlock O && 90.83 & 88.96 & 90.39 && 74.91 & 71.90 & 72.28 \\
& \faUnlock E \faUnlock B \faUnlock O \faExchange A && \bf 91.08 & \bf 90.73 & \bf 90.99 && \bf 75.38 & \bf 75.61 & \bf 75.43 \\
& \faSpinner E \faSpinner B \faSpinner O  && 89.66 &  90.20 & 89.79 && 67.45 & 72.66 & 68.55 \\

\hline
\end{tabular}

\caption{\small \label{tab:main-result} \small Performance of  the source model ($\mathcal{M}_{\text{S}}$) and the target model ($\mathcal{M}_{\text{T}}$), according to different settings. The reported performance is the F1 score on the test set.
Ori. indicates the original 3 NE categories in the source data, while New indicates the new NE categories in the target data. All is the overall test F1 in the subsequent step (for all 4 NE categories).
}
\end{table*}
 
Secondly,  we explored several settings of updating the model weights in the subsequent step.
 The performance are shown in Table \ref{tab:main-result}:
 E, B, O represent the weight parameters in the Embedding, the Bidirectional LSTM, and the Output layers, respectively.
 The parameters associated with CRF are included in the notation of \say{O}.
 Our settings include: (1) the weights of layers before the output layer are fixed, and the weights of the output layer are updated (\faLock{E}\faLock{B}\faUnlock{O}); (2) the weights of layers before the output layer are fixed, the transferred part of the output layer is fixed too, while the rest is updated (\faLock{E}\faLock{B}\faAdjust{O}); (3) none of the weights in the model is fixed (\faUnlock{E}\faUnlock{B}\faUnlock{O}); (4) as a baseline, all the weights are not transferred but randomly initialized (\faSpinner{E}\faSpinner{B}\faSpinner{O}). Note that in the setting (1), (2) and (3), the model weights are initialized with those from the source model.
 
 \subsubsection{Performance of Transferring weights}
 Compared to randomly initializing the model weights (setting \faSpinner E\faSpinner B\faSpinner O), in the transfer learning step, we find out that regardless of whether the weights are fixed or not, transferring weights from the initial model always boosts the performance.
 On CONLL dataset, transferring model weights improves the performance ranging from 0.60 to 3.50 points, while on I-CAB dataset, it gives 7.32 points of increment in F1.
 In order to verify that the model with transferred parameters is indeed better on the target task (not just because it has faster convergence rate), we further carried out additional training using a larger number of epochs for the baseline models with randomly initialized parameters. 
 Even in this condition, the baseline models still did not produce a better performance than what is reported in the table. 
This suggests that the better results come from the transferred parameters.

\subsubsection{Update of Model Parameters}
 Though transferring learned weights is helpful in reaching a better performance, keeping the learned weights fixed produces worse results, especially for the new NE category.
 On both datasets, the \faLock{E}\faLock{B}\faUnlock{O} and \faLock{E}\faLock{B}\faAdjust{O} settings perform poorly in recognizing the new NEs. 
 The results are also worse than setting \faUnlock{E}\faUnlock{B}\faUnlock{O} for both BLSTM and BLSTM$+$CRF model.
 
  In the standard pre-training and fine-tuning TL paradigm, usually only the output layer is fine-tuned on the target data.
 We argue this is not the best setting for our experiment because of two reasons:
 firstly, in our target data, there are still NEs that appeared in the source data, hence there is information to be used to further train parameters in the model with regard to these NE entities. 
 Secondly, the knowledge learned about the \texttt{null} label is adverse to the recognition of new NE labels.
 The progressive NER is a TL scenario without crossing domain. The source and target domains share a high similarity.
 The source and target tasks differ only in the output space. 
 Hence updating all the model parameters provides more benefits rather than causing catastrophic forgetting.
 It also helps the model to recover from the labeling disagreement of new NE in the source and target data.
 In fact,  \faUnlock{E}\faUnlock{B}\faUnlock{O}, which can fine-tune all the weights, achieved the best performance, for both models on both datasets, compared to other parameter update settings.

  \begin{figure*}[!ht]
    \centering
        \subfloat[10 \%]{\includegraphics[width=0.45\columnwidth]{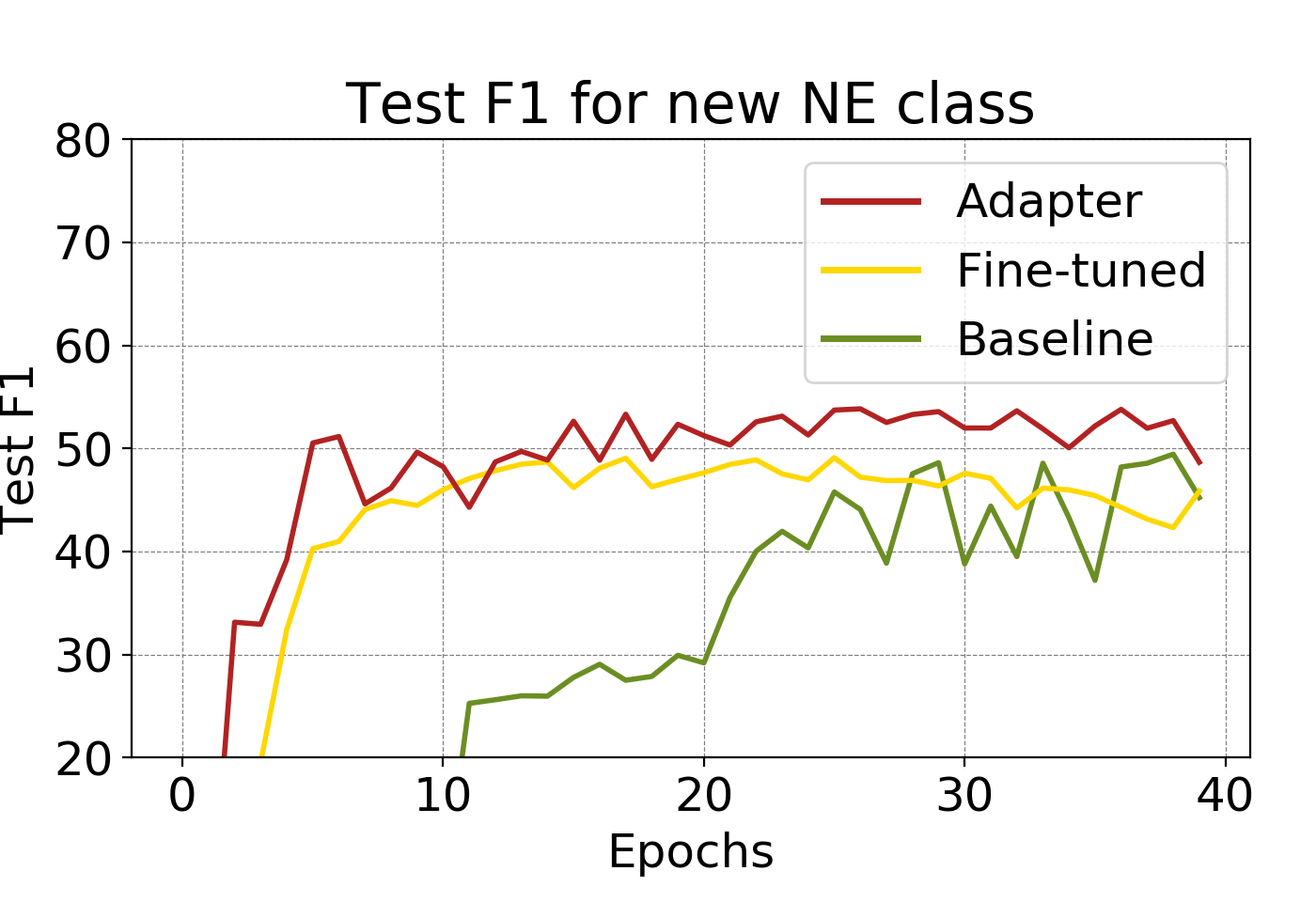}}
        \subfloat[25 \%]{\includegraphics[width=0.45\columnwidth]{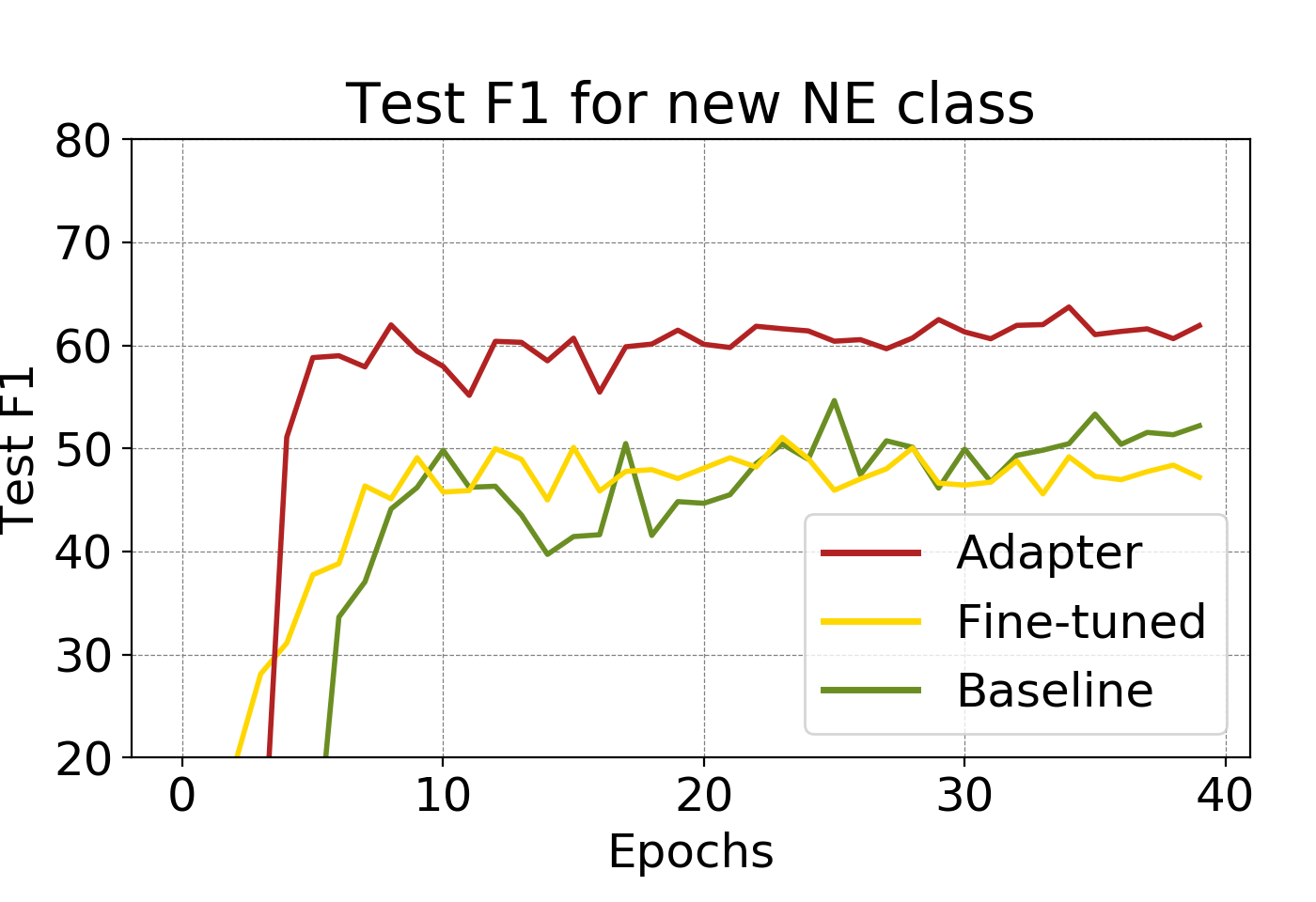}}
        \subfloat[50 \%]{\includegraphics[width=0.45\columnwidth]{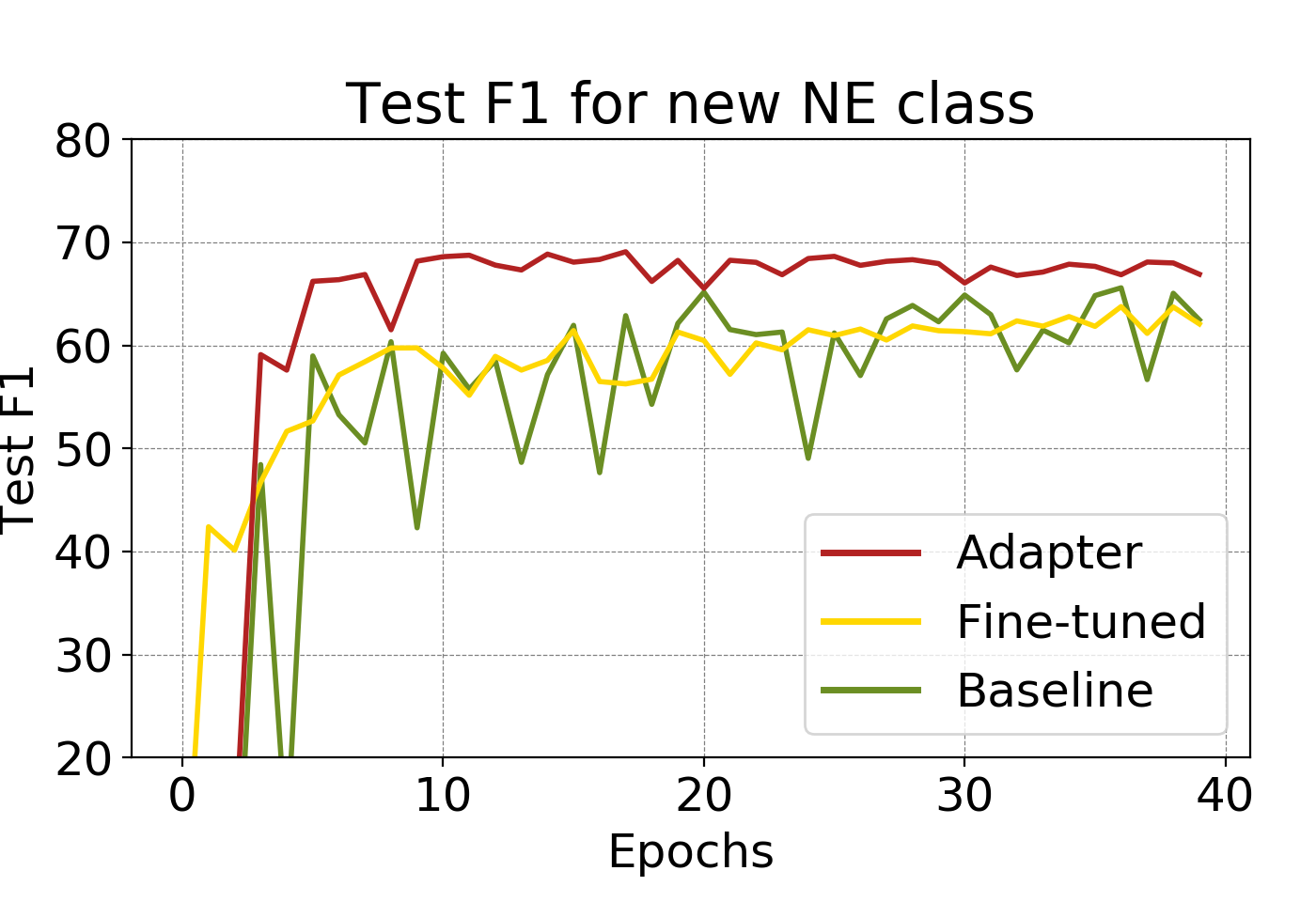}}
        \subfloat[100\%]{\includegraphics[width=0.45\columnwidth]{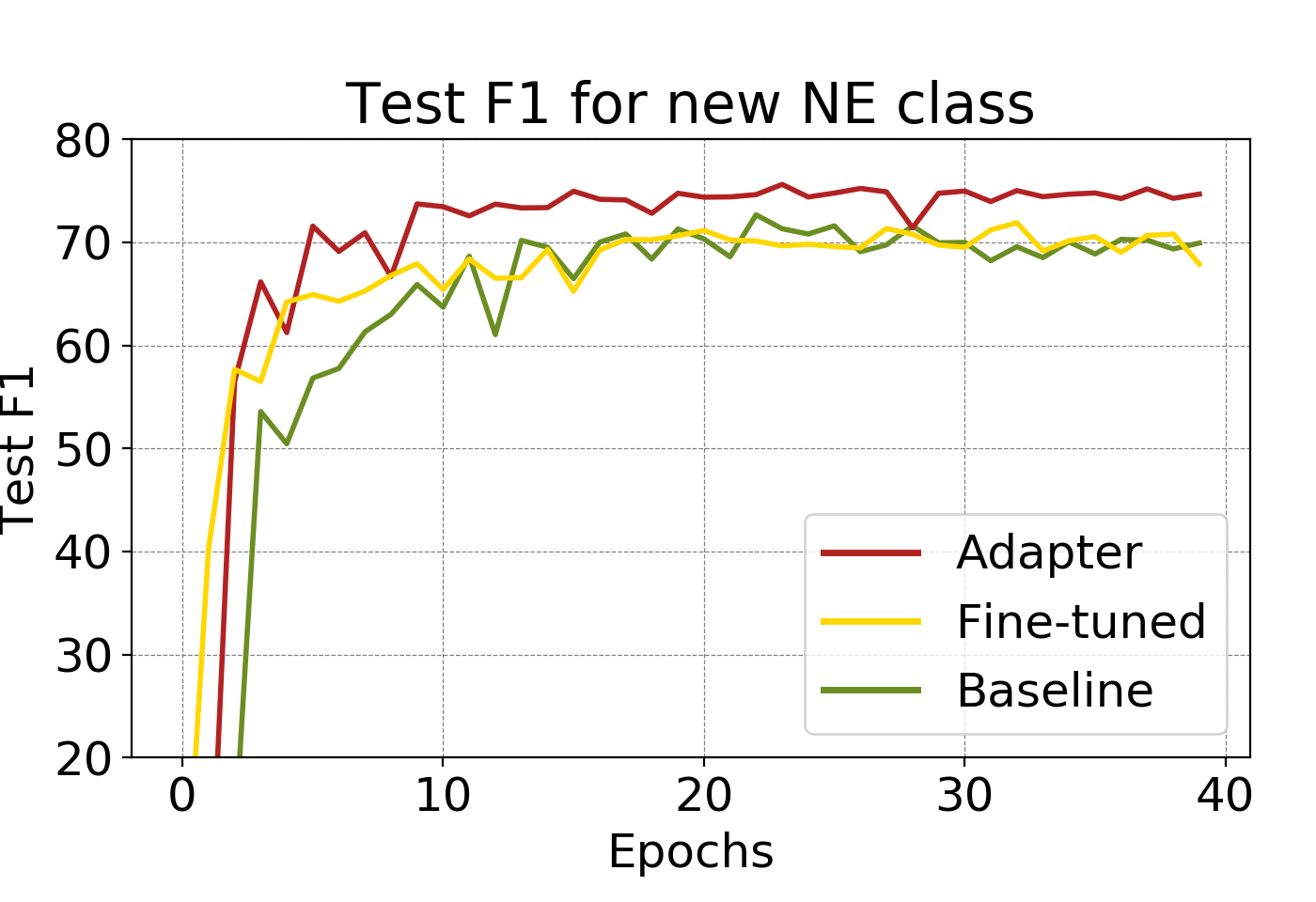}}
    \caption{\small Model F1s evaluated on the test set varying the size of the available training target domain data}
    \label{fig:increment-plot}
\end{figure*}

 \subsubsection{Improvement by Using the Adapter}   
 We further analyze the results with regard to using the adapter. 
 The overall F1 score of both BLSTM and BLSTM$+$CRF models (with setting \faUnlock{E}\faUnlock{B}\faUnlock{O}) suffers from a certain amount of degradation on the target domain.
This happens for both the original three NE categories and the new NE category.

The  comparison between the results of the transferred models with adapter (\faUnlock{E}\faUnlock{B}\faUnlock{O}\faExchange{A}) and those without adapter (\faUnlock{E}\faUnlock{B}\faUnlock{O}) shows a consistent improvement on F1 score over the original NE categories.
 In some cases, for example, while using the adapter on the I-CAB dataset, the transfer model performance of the original NE categories even surpasses the F1 of the source model.
 It suggests that the adapter manages to mitigate the knowledge forgetting, and enabling the model to fine-tune on these original NEs.
 
\begin{figure}[!t]
    \centering
        \subfloat[New category (LOC), CONLL'03]{\includegraphics[width=0.40\columnwidth]{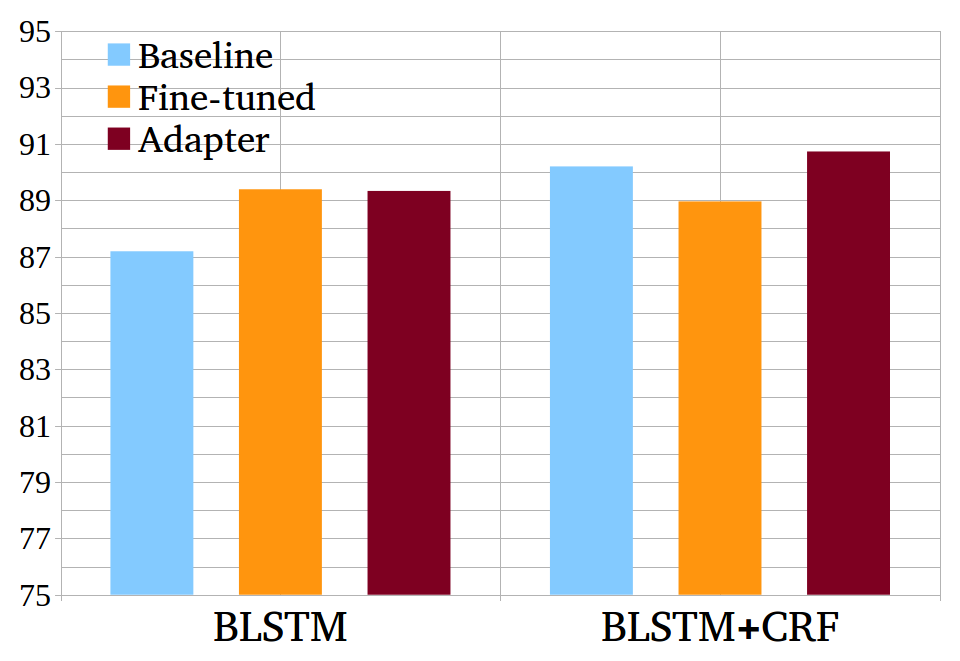}}
        \qquad
        \subfloat[Overall, CONLL'03]{\includegraphics[width=0.40\columnwidth]{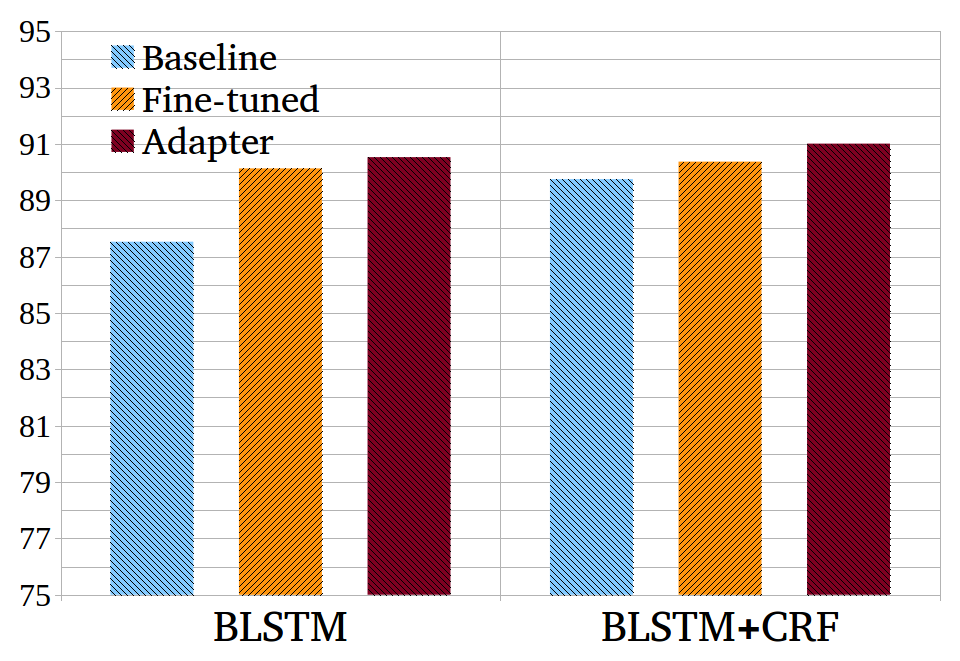}}
\\
        \subfloat[New category (GPE), I-CAB]{\includegraphics[width=0.40\columnwidth]{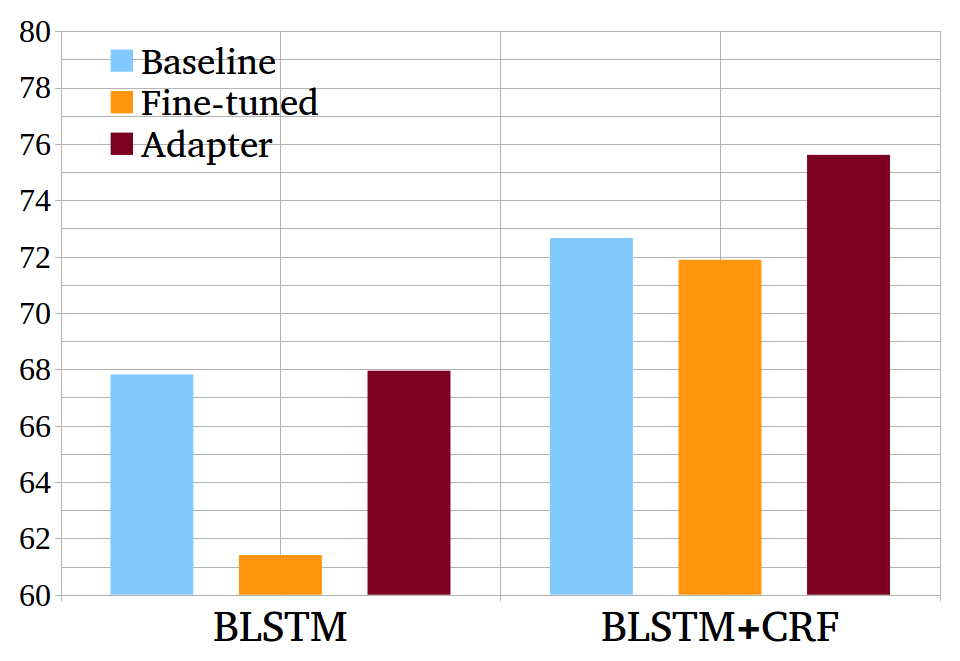}}
        \qquad
        \subfloat[Overall, I-CAB]{\includegraphics[width=0.40\columnwidth]{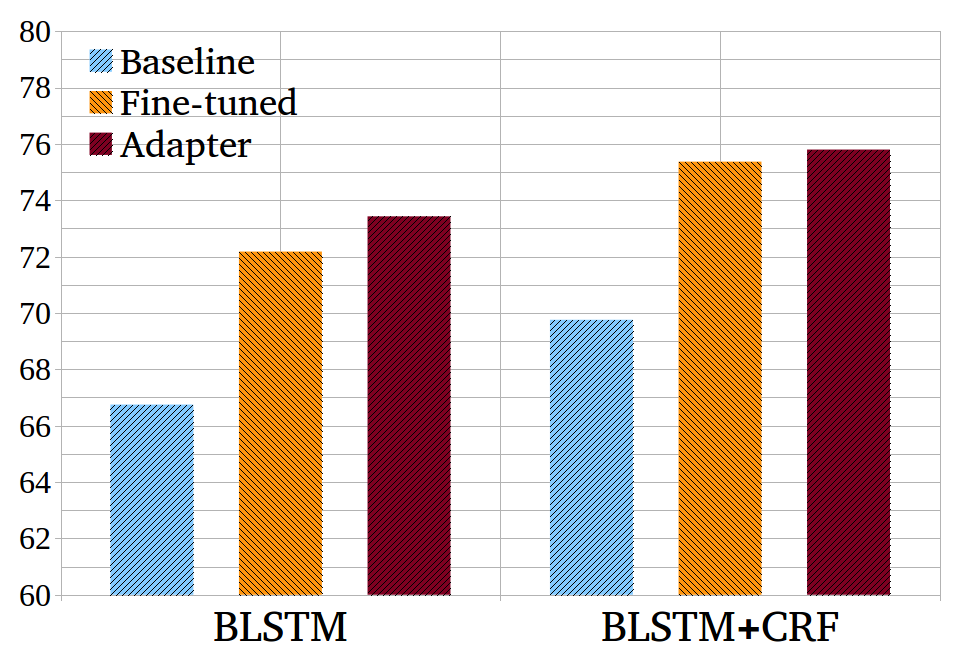}}
        
    \caption{\small Overall and single category Test F1 of baseline model (\faSpinner{E}\faSpinner{B}\faSpinner{O}), fine-tuned model (\faUnlock{E}\faUnlock{B}\faUnlock{O}), and model with adapter (\faUnlock{E}\faUnlock{B}\faUnlock{O}\faExchange{A}) on CONLL and I-CAB }
    \label{fig:adapter-result}
\end{figure}

As for the new NE category, in almost all cases, transferring with adapter helps to better recognize them. We show in detail the improvement obtained by using the adapter in Figure~\ref{fig:adapter-result}. 
 It is worth noting that on I-CAB, the adapter produces an improvement of F1 on the new NE for both BLSTM and BLSTM$+$CRF models (6.54 and 3.71 points respectively). 
 The improvement is also observed in the results of CONLL dataset. This indicates that the adapter is able to help in resolving the annotation disagreement between the source and the target data. The improvement is less obvious on the CONLL dataset because the NEs are fairly easy to learn for both BLSTM and BLSTM+CRF model. Indeed, with a small amount of training data (e.g., the baseline setting), the F1 is already rather good. Instead, on I-CAB, there is more headroom, thus the adapter can have a larger impact.
 
In Figure \ref{fig:increment-plot}, we show the F1s of models evaluated on the test set, varying the size of the available training data in the target domain. This is useful to analyze whether the adapter is helpful in the situation of smaller amount of labeled data for new NE.
 In each sub-level figures, the lines represent F1 scores of
the BLSTM$+$CRF baseline model (\faSpinner{E}\faSpinner{B}\faSpinner{O}), the fine-tuned model (\faUnlock{E}\faUnlock{B}\faUnlock{O}), and the model with adapter (\faUnlock{E}\faUnlock{B}\faUnlock{O}\faExchange{A}) according to the increasing number of epochs.
 The plot shows that using the adapter consistently helps to recognize the new NE, especially when small training data is available in the target domain (only 10\% or 25\%) and the baseline and transferred models both show difficulties in learning the new NE category. 
 The plots also show that the adapter makes learning smoother.
 
 We finally observe that the transferred models with neural adapter converge faster during the subsequent training (around 20 epochs) in comparison to other models. This further suggests that the transferred models are able to learn to predict new categories rapidly in a shorter time.

\subsection{Results of all NE categories}
It is important to ensure that the improvement in the performance is not specific to a target NE category.
Thus, we performed additional experiments on CONLL and I-CAB dataset, using other NEs as the target in the subsequent step.

In Table~\ref{tab:other_result}, we present the overall F1 scores obtained by BLSTM$+$CRF model while recognizing different new NE categories.
The first column in the table identifies different target NE categories. The other three columns present the results of the models without any TL method (Baseline), with the transferred parameters (W/o Adapter), and with the neural adapter (W/ Adapter), respectively.
The results indicate a consistent improvement in the F1 score using transferred parameters and especially using the neural adapter.
On average, our best TL model with neural adapter gains 1.83 points of improvement in F1 score on CONLL dataset, and 10.30 points on I-CAB, compared to that obtained by the baseline model.
It is evident that our proposed methods are able to improve the performance of recognizing NEs in the target data, regardless of dataset or target NE types.

\begin{table}[t]
\centering
\small
\begin{tabular}{C{0.8cm}|C{1.6cm}|C{1.7cm}|C{1.9cm}}
\hline
\multicolumn{4}{c}{\textbf{CONLL 2003}} \\
\hline
 & Baseline (\faSpinner{E}\faSpinner{B}\faSpinner{O}) & W/o Adapter (\faUnlock{E}\faUnlock{B}\faUnlock{O})& W/ Adapter (\faUnlock{E}\faUnlock{B}\faUnlock{O}\faExchange{A})\\ 
\hline
LOC&89.79&90.39&90.99\\\hline
PER&88.33&90.23&90.36\\\hline
ORG&88.77&89.28&90.16\\\hline
MISC&87.64&90.30&90.34\\\hline
\hline
\multicolumn{4}{c}{\textbf{I-CAB 2006}} \\
\hline
LOC & 64.39 & 75.49 & 76.87 \\ \hline 
PER & 59.98 & 70.74 & 72.82 \\ \hline
ORG & 64.65 & 72.64 & 73.63 \\ \hline
GPE & 68.55 & 72.28 & 75.43 \\ \hline
\end{tabular} 
\caption{\small \label{tab:other_result} Overall F1 score in recognizing different target NE categories of the test set of the subsequent step}
\end{table}

\section{Conclusion}
In this paper, we have studies TL for sequence labeling tasks. In  particular, we experiment with a progressive NER setting, simulating real-world applications of NER. We verified that our methods can be applied to current state-of-the-art neural models for NER. We proposed a neural adapter for connecting the target and the source models to mitigate the forgetting of learned knowledge. 
 We carried out extensive experiments to analyze (i) the effect on the performance of the transfer approach,
(ii) how the parameters in the transferred model should be initialized and (iii) how the parameters should be updated. 
The empirical results show the effectiveness of the proposed methods and techniques. We will make data and models available to support this new line of research.
In future work, we would like to test our approach to several different sequence labeling tasks to fully demonstrate the generality of our approach.
 
 \section*{Acknowledgements}
This research was partially supported by Almawave S.r.l. We would like to thank  Giuseppe Castellucci, Andrea Favalli, and Raniero Romagnoli for inspiring this work with useful discussions on neural models for applications to real-world problems in the industrial world.

\bibliography{aaai2019}
\bibliographystyle{aaai}

\end{document}